\documentclass[a4paper,fleqn]{cas-dc}

\usepackage[numbers]{natbib}

\def\tsc#1{\csdef{#1}{\textsc{\lowercase{#1}}\xspace}}
\tsc{WGM}
\tsc{QE}
\tsc{EP}
\tsc{PMS}
\tsc{BEC}
\tsc{DE}

\begin{document}
\let\WriteBookmarks\relax
\def\floatpagepagefraction{1}
\def\textpagefraction{.001}

\shorttitle{AGI for Agriculture}

\shortauthors{Lu et~al.}

\title [mode = title]{AGI for Agriculture}

\author[1]{Guoyu Lu }[]\ead{guoyu.lu@uga.edu}
\fnmark[1]

\author[2, 1]{{Sheng Li} }\ead{shengli@virginia.edu }
\fnmark[1]

\author[1]{{Gengchen Mai} }\ead{gengchen.mai25@uga.edu}

\author[1]{{Jin Sun} }\ead{jinsun@uga.edu}

\author[4]{{Dajiang Zhu} }\ead{dajiang.zhu@uta.edu}

\author[1]{{Lilong Chai} }\ead{lchai@uga.edu}

\author[1]{{Haijian Sun} }\ead{hsun@uga.edu}

\author[1]{{Xianqiao Wang} }\ead{xqwang@uga.edu}

\author[1]{{Haixing Dai} }\ead{haixing.dai@uga.edu}

\author[1]{{Ninghao Liu} }\ead{ninghao.liu@uga.edu}

\author[3]{{Rui Xu} }\ead{asilan@ufl.edu}

\author[3]{{Daniel Petti} }\ead{daniel.petti@ufl.edu}

\author[3]{{Changying Li} }\ead{cli2@ufl.edu}
\cormark[1]

\author[1]{{Tianming Liu} }\ead{tianming.liu@gmail.com}
\cormark[1]

\affiliation[1]{organization={University of Georgia},
city={Athens},
postcode={30602}, 
    state={GA},
    country={USA}}
\affiliation[2]{organization={University of Virginia},
city={Charlottesville},
postcode={22903}, 
    state={VA},
    country={USA}}
\affiliation[3]{organization={University of Florida},
city={Gainesville},
postcode={32611}, 
    state={FL},
    country={USA}}
\affiliation[4]{organization={University of Texas at Arlington},
city={Arlington},
postcode={76010}, 
    state={TX},
    country={USA}}

\cortext[cor1]{Corresponding author}

\fntext[fn1]{These authors contributed equally to this work.}

\begin{abstract}
Artificial General Intelligence (AGI) is poised to revolutionize a variety of sectors, including healthcare, finance, transportation, and education. Within healthcare, AGI is being utilized to analyze clinical medical notes, recognize patterns in patient data, and aid in patient management. Agriculture is another critical sector that impacts the lives of individuals worldwide. It serves as a foundation for providing food, fiber, and fuel, yet faces several challenges, such as climate change, soil degradation, water scarcity, and food security. AGI has the potential to tackle these issues by enhancing crop yields, reducing waste, and promoting sustainable farming practices. It can also help farmers make informed decisions by leveraging real-time data, leading to more efficient and effective farm management.
This paper delves into the potential future applications of AGI in agriculture, such as agriculture image processing, natural language processing (NLP), robotics, knowledge graphs, and infrastructure, and their impact on precision livestock and precision crops. By leveraging the power of AGI, these emerging technologies can provide farmers with actionable insights, allowing for optimized decision-making and increased productivity. The transformative potential of AGI in agriculture is vast, and this paper aims to highlight its potential to revolutionize the industry.
\end{abstract}

\begin{keywords}
 Artificial General Intelligence (AGI) \sep Agriculture
\end{keywords}

\maketitle

\section{Introduction}

Since the artificial intelligence application AlphaGo defeated the world Go masters, artificial intelligence has become a hot topic again, particularly after the release of ChatGPT and GPT-4 ~\citep{OpenAI2023}. Artificial general intelligence (AGI) is that machines can think like humans and perform a multitude of general tasks by means of transfer learning and various other modalities ~\citep{zhao2023brain}. Over the past few years, Large Language Models (LLMs) \citep{holmes2023evaluating} such as GPT-3.5 used in ChatGPT have attracted a lot of attention and interest. More recently, GPT-4 is a multimodal language and vision system with the ability to recognize and describe pictures. Once launched, it impresses everyone with its ability to match human intelligence. With further development, AGI is expected to have deeper and broader impacts over agriculture. 

Despite agriculture being recognized as a fundamental pillar for national and global economic development, the scientific literature on the implementation of Artificial Intelligence (AI), particularly Artificial General Intelligence (AGI), in agriculture remains relatively scant. This underrepresentation in the literature is striking given the importance of agriculture as a crucial component of global food security and the potential for AI to enhance agricultural productivity and sustainability. With the recent fast development of AGI, we expect it will play an increasingly critical role in the future agriculture applications. Therefore, in this article, we will discuss the applications of AGI in agriculture.

\begin{figure*}[t] 
\centering
\includegraphics[width=\textwidth]{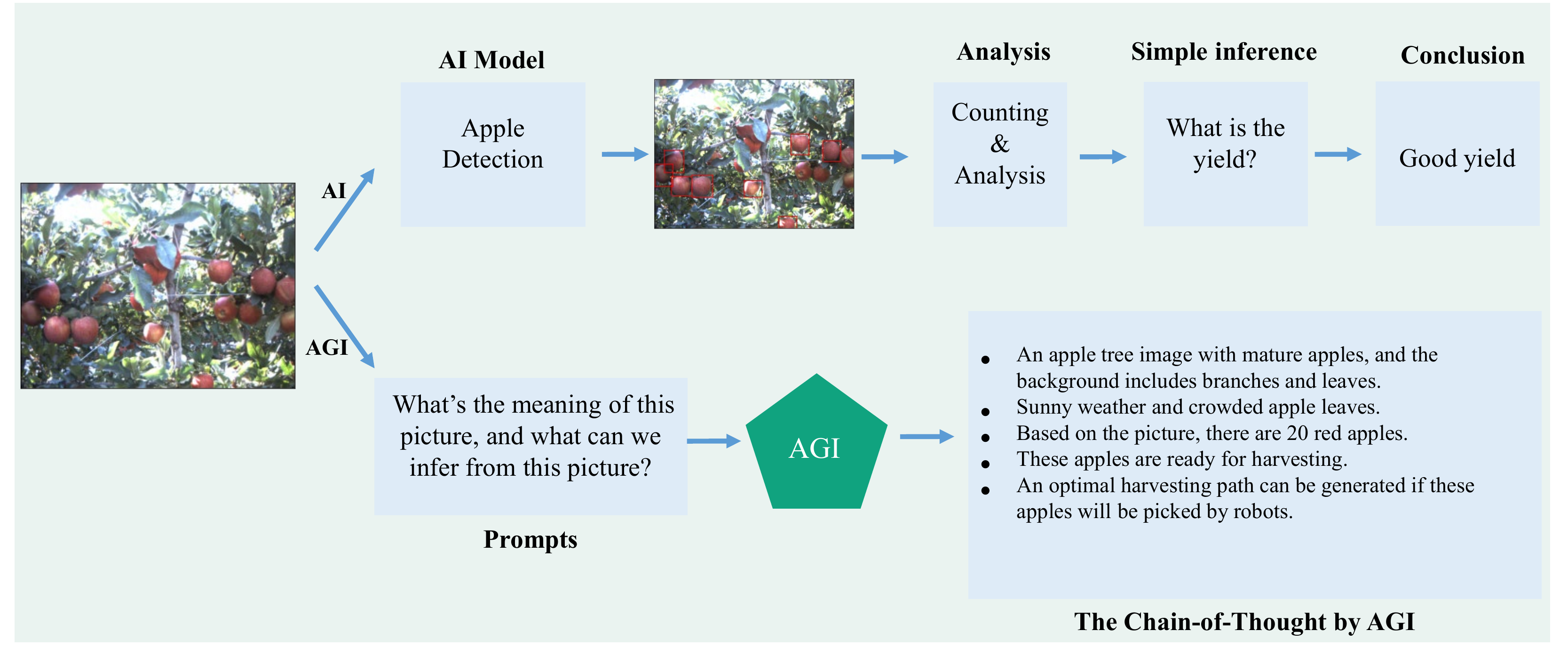} 
\caption{Main pipeline of AGI for agriculture applications in comparison with specific AI-CV-based agriculture image processing framework.} 
\label{agipipeline}
\end{figure*}

The application of computer vision mainly includes target detection \citep{redmon2016yolo}, image segmentation \citep{kirillov2023segany, he2017maskrcnn}, localization \citep{lu2022image, lu2015localize}, and image classification \citep{he2016resnet}.  AGI technology can apply computer vision technology to deeper and wider fields. In the field of agriculture, AGI technology can also enable and promote the development of a wide range of applications, such as crop pest identification, disease detection, etc. In addition, AGI can also be used in precision farming, which involves the use of advanced sensors and algorithms to analyze data and optimize crop yields. This AGI technology can help farmers make informed decisions about irrigation, fertilization, and other farming practices, leading to increased productivity and efficiency. AGI can also be used in developing new crop varieties through computational breeding, which could speed up the process of developing more resilient, high-yield crops. Another potential application of AGI in agriculture is in farm automation, where robots and drones equipped with advanced computer vision and machine learning capabilities could perform a wide range of tasks, from planting and harvesting to monitoring crops and livestock. \textbf{In comparison with major computer vision or other specific AI (i.e., artificial narrow intelligence (ANI) \citep{kuusi2022scenarios}) based methods, AGI could potentially provide more comprehensive, smarter, and richer information with fewer middle steps}, as illustrated in Fig. \ref{agipipeline}.

Notably, AGI could support agriculture development from multiple modalities, such as images, sound, robotics, knowledge graph, NLP, etc., and potentially fuse them to support decision-making. One significant difference between AGI and AI in agriculture is their ability to generalize and learn from new situations. AGI systems have the ability to transfer learning from one domain to another and adapt to new situations without human intervention. This means that AGI can learn from new data and experiences in a way that AI cannot. \textbf{On the one hand, AGI could enable extensive new applications; on the other hand, AGI can significantly enhance the performance of existing AI systems on the same tasks.} 

AGI has the potential to revolutionize the way farmers interact with agriculture robots. Using natural language processing and machine learning, AGI can enable robots to understand and respond to human commands, whether they are given through voice or text. This will help simplify the operation of agriculture robots, reduce the learning time for new users, and make them more accessible to farmers. One potential application of AGI in agriculture is the development of smart farming assistants that can perform tasks such as planting, watering, and harvesting crops in response to natural language commands. These assistants can use machine learning algorithms to optimize their recommendations based on real-time data from the farm and adapt to individual farmer's preference.

AGI can also facilitate the coordination and cooperation of robots to optimize farm operations. For example, robots equipped with AGI can communicate with each other to prioritize tasks, optimize resource utilization, and avoid collisions and other accidents. By enabling robots to work together more efficiently, AGI can help farmers increase their productivity, reduce costs, and make better use of their resources.

For breeding and phenotyping, by analyzing large volumes of phenomic and genomic data and other environmental factors, AGI can help breeders to identify and select the most promising plant or animal traits with greater precision and efficiency than traditional methods. Based on breeders' needs, AGI can possibly generate predictive models that can forecast the performance of different breeding combinations based on phenomic and genomic data and other relevant factors. These models can use machine learning algorithms to optimize breeding strategies and predict the outcome of different breeding combinations.

More apparent, broader, and quicker impacts that AGI can bring to agriculture are its capability to enhance the effectiveness of existing AI systems. An example is the identification of types of crop pests and diseases. Crop diseases and insect pests are one of the main disasters in agriculture. They have the characteristics of various types, great impact, and frequent outbreaks. At present, in order to control the occurrence of pests and diseases, most farmers spray pesticides blindly, which inevitably causes a series of problems such as environmental pollution and food safety \citep{pesticide, tudi2021agriculture}. In addition, the identification of crop diseases and insect pests mainly depends on the experience of farmers and experts. When dealing with large-scale monitoring of crop diseases and insect pests, current AI systems need to rely on data that have already been well-labeled to train comprehensively. It is easy to miss the rescue measures and the best time for diseased crops. AGI technology can lower the threshold for identifying types of crop diseases and insect pests. Farmers can take photos of plant diseases by their mobile phones, and then upload them to the AGI model. AGI will be able to understand the input instructions and tell farmers how to manage identified pests and diseases, e.g., which pesticide to buy. In addition, farmers can also capture the sound of pests in the field by recording equipments and upload them to the AGI model. AGI can then recognize the sound characteristics of pests and perform speech recognition to identify crop pests. In contrast with normal AI and CV algorithms, most notably, AGI can help farmers identify previously unknown pests and diseases by learning from new data and recognizing patterns that may not be apparent to humans. This ability to generalize and learn from new situations could significantly enhance the effectiveness of agriculture solutions and improve the overall resilience of the industry.

In another scenario, weeds are similar in color to crops, and their growth also requires sunlight and water resources. Weeds will invade the living space of crops and rob the crops of nutrients. If the density of weeds in the field is too high, it will seriously affect the yield and quality of crops. Therefore, weeding is an inevitable and important part of agricultural production. AGI technology can now help people get rid of weeds. AGI has the potential to outperform traditional AI systems. AGI can be trained to identify and differentiate between plants and weeds, and then selectively remove weeds without harming the crops. This is possible through the use of advanced computer vision and machine learning algorithms that allow AGI to recognize and respond to new situations and data. On the one hand, through the general-purpose vision and reasoning capabilities in AGI technology, people can accurately locate weeds, guide people to quickly reach the weed growth area, and inform people of suitable pesticides, so that people can have different responses to different weeds, and make targeted weed control. Traditional AI systems for weeding typically rely on pre-programmed rules and limited data sets, which may not be able to adapt to new situations or recognize emerging weed species. AGI systems can learn from their experiences and adapt to new data, allowing them to become more accurate and efficient over time. Another way is to connect AGI to intelligent robots, so that the robots can quickly and accurately identify weeds and spray pesticides precisely. This will reduce the use of pesticides such as herbicides, reduce environmental pollution caused by pesticides, and avoid labor shortages, further reducing costs.

The quality of crop seeds not only determines the survival rate of seedlings, but also has an important impact on the harvest and quality of crops. High-vigorous seeds have strong resistance to stress. When planted in the field, they have a high germination rate, rapid and neat emergence, and can produce high-yield and high-quality crops; low-vigority seeds have poor stress resistance, and are prone to rotten seeds and dead seedlings, resulting in crop yield reduction. AGI can screen unknown seeds by extracting excellent seed traits and their corresponding data, and retain seeds with excellent shapes. AGI can further support seed selection based on multiple clues, such as their size, plumpness, external geometric characteristics, color characteristics, and whether there is mildew, etc.

Fruit grading is to strictly screen the fruits before entering the market, remove rotten fruits, and classify them according to the quality of the fruits. Through the detection of fruit photos, AGI can package, transport and sell different grades of fruits separately, reduce storage losses, and discuss prices according to quality when selling to obtain maximum economic benefits. AGI has a significant advantage over traditional AI in its ability to generalize and learn from new situations. Therefore, AGI can be trained to recognize various fruit quality factors such as ripeness, size, color, and shape, and accurately sort fruits into different quality categories such as "high quality," "average," and "not suitable to sell." In contrast, when fruit shape, size, color, and especially rotten regions and degrees differ from limited training data, the screening effect may be reduced. 

Furthermore, AGI has the potential to perform a wide range of tasks in agriculture autonomously. With its ability to understand and interpret natural language and visual cues, AGI can make decisions and take actions independently without human intervention. This level of autonomy is not possible with current AI systems and could revolutionize the way farmers manage their crops and livestock.

With the examples mentioned above, AGI could provide critical support and service for agriculture development based on information extracted from multiple modalities and knowledge domains. The following sections of this paper will articulate AGI for agricultural applications from multiple perspectives in detail, including AGI background, agriculture image understanding, NLP, knowledge graph, infrastructure, robotics, and their applications to precision crops and precision livestock, respectively.

 \section{Background}

In this section, we briefly introduce how AI and AGI techniques, such as natural language processing (NLP), can be adopted to advance a wide variety of agriculture applications. 
Texts are probably the most accessible data sources in many domains including agriculture, considering the vast amount of documents (e.g., scientific papers, technical reports, surveys, news, and social media data). Although some pioneering efforts use NLP techniques for agriculture applications, customizing NLP models for specific agriculture applications has been largely underexplored. In the following, we discuss several types of NLP techniques, including both traditional NLP techniques and the most recent large language models (LLMs), which have been used (or could be potentially used) to assist agriculture applications.

\textbf{Information Extraction.} As a fundamental task in NLP, information extraction focuses on extracting structured information from unstructured or semi-structured documents. It is also an essential step for many other NLP tasks such as knowledge graph construction~\citep{rezayi2021edge}. Some recent efforts in the agriculture domain can be found in this survey paper~\citep{nismi2022review}. Most work in this area focuses on applying classical NLP techniques (e.g., rule-based methods and learning-based methods) for subtasks like named entity recognition (NER) and then creating ontology or knowledge graph that is specific to agriculture. Recently, transformer-based language models like BERT have also been used for entity and relation extraction~\citep{qiao2022joint}. 

\textbf{Semantic Matching.} Considering the diversity in language representations, semantic matching is a key technique that enables the correct matching of semantically similar entities with varying lengths (e.g., words, phrases, sentences, documents). One recent application of semantic matching in the agriculture domain is to establish the mapping between food descriptions and nutrition database~\citep{rezayi2022agribert}. In particular, a food-related text corpora (e.g., food and agricultural literature) is collected for pretraining transformer-based language models. And semantic matching between food and nutrition is then performed in the embedding space, by formulating an answer selection problem. Moreover, semantic matching could play an essential role in fusing agricultural text data from different sources.    

\textbf{Question Answering and Dialogue System.} Question answering (QA) and dialogue systems are fundamental problems in NLP~\citep{chen2017survey,lai2018review}. The models are expected to accurately understand the question and context, and then generate informative responses. A direct extension of the QA technique is dialogue systems such as chatbots. By far, some chatbots have been developed in the agriculture domain, such as answering questions from farmers~\citep{niranjan2019survey,sukumar2021text,gounder2021agrobot}. For instance, a recurrent neural network (RNN) based chatbot is designed to deal with questions about soil testing, plant protection, and nutrient management~\citep{sukumar2021text}. These QA systems or chatbots, however, have very limited capability due to the small model size and insufficient training data.

\textbf{Pretrained Language Models (PLMs).} Language modeling has become the mainstream technique in NLP due to its effectiveness and remarkable generalizability. In particular, transformer-based pretrained language models can serve as base models for lots of NLP applications. PLMs such as BERT are usually pretrained by tasks like masked language modeling (MLM) and next sentence prediction (NSP). Since these PLMs are pretrained in general domains using data from Wikipedia and BookCorpus, a fine-tuning step is usually required for specific domains. By fine-tuning the base PLMs with high-quality text corpus from a specific domain, customized language models could be developed. Some representative examples include BioBERT~\citep{lee2020biobert}, PubMedBERT~\citep{gu2021domain}, and LEGAL-BERT~\citep{chalkidis2020legal}. In~\citep{rezayi2022agribert}, a large-scale corpus of agricultural literature with more than 300 million tokens is used to fine-tune the generic BERT into AgriBERT. Such domain-specific language models could be potentially used to assist many agricultural applications, such as agricultural marketing, agricultural social media analytics, etc.   

\textbf{Large Language Models (LLMs).} LLMs usually refer to PLMs with more than billions of model parameters, such as the GPT (Generative Pre-trained Transformer) models (GPT-1~\citep{radford2018improving}, GPT-2~\citep{radford2019language}, GPT-3~\citep{brown2020language}, InstructGPT~\citep{ouyang2022training}, ChatGPT, and GPT-4~\citep{OpenAI2023}), LLaMA (Large Language Model Meta AI)~\citep{touvron2023llama}, and BLOOM~\citep{scao2022bloom}. For instance, GPT-3 and BLOOM have 175 billion parameters, and LLaMA has 65 billion parameters. Similar to traditional PLMs like BERT, these recent LLMs also follow the Transformer architecture~\citep{vaswani2017attention}, but they differ from each other in training objectives, data sources, optimization details, etc. For example, BERT only adopts the encoder of Transformer as it focuses on learning language representations, while GPT models adopt the decoder of Transformer, as their major objective is to generate realistic text. Owing to their huge model sizes and effective incorporation of human feedback (e.g., prompts), the latest LLMs like ChatGPT and GPT-4 have not only achieved amazing performance in traditional NLP tasks like language understanding, language generation, and question answering, but also demonstrated remarkable capabilities in zero-shot learning and reasoning~\citep{liu2023summary}. For instance, ChatGPT has shown a very promising capability in inductive reasoning and deductive reasoning~\citep{holmes2023evaluating}. GPT-4 is a large multimodal model that can deal with both language and visual inputs. In addition, LLMs are also effective tools for data augmentation~\citep{dai2023chataug}, which is very helpful for certain real-world scenarios with limited data. These exciting observations showcase that the LLMs-based techniques may enable a promising roadmap towards AGI. For agricultural applications, the LLMs would be able to significantly advance the processing, understanding, and interpretation of agricultural literature and documents. Moreover, the zero-shot learning and reasoning capabilities of LLMs would also greatly facilitate various decision-making problems in the agriculture domain. 

In the following sections, we will discuss how the AGI techniques (such as LLMs and large multimodal models) could be leveraged to advance a broad range of research areas and applications in agriculture. Sections 3-7 focus on several research topics in agriculture, involving different types of data (e.g., images, knowledge graphs) and various applications (e.g., precision livestock and crops, agricultural robotics). In Section \ref{AGI-Image}, we will introduce AGI for agriculture image understanding, such as managing metadata in breeding studies, generating images for data augmentation, multi-task scheduling and compositing, few-shot learning, and domain adaptation. Section \ref{AGI-KG} discusses several potential aspects where knowledge graph techniques and AGI are jointly exploited for agricultural applications. Section \ref{AGI-Robotics} describes several possible ways of using AGI for agriculture robotics. Section \ref{AGI-application} presents a few case studies, including AGI for precision farming and phenomics, introduces, AGI for precision livestock, and AGI for agricultural infrastructure. The last section concludes this paper.  \section{AGI for Agriculture Image Understanding}
\label{AGI-Image}

\subsection{Generative AI for Agricultural Data}

Recently, generative AI models such as the Stable Diffusion~\citep{stablediffusion} model have shown impressive high-quality generation of photorealistic images and videos \citep{ho2022imagenvideo}.
Similar breakthroughs have occurred in other domains such as speech \citep{whisper}, all powered by large-scale generative AI models.
Multi-modal LLMs are inherently generative models \citep{deeplearningbook}. They can be fine-tuned to capture the complex distribution of rich agriculture data in satellite images, videos, hyperspectral sensing data, 3D point clouds, as well as time-series measures. Such generative models can empower the following tasks:

\smallskip
\noindent\textbf{Training data generation}. One limitation of applying specialized computer vision algorithms such as PointNet \citep{qi2016pointnet} or MaskRCNN \citep{he2017maskrcnn} to agricultural visual data is the scarcity of training data and labels. Obtaining high-quality data are time-consuming and labeling them properly is even more expensive \citep{zhou2017bigdata}. One possible solution is to build a digital environment that generates synthetic data\footnote{\url{https://www.nvidia.com/en-us/omniverse/}} but there is usually a domain gap between the simulated environment and the real world. Complex solutions such as domain adaptation \citep{csurka2017domainsurvey} have to be applied to address this issue. On the other hand, multi-modal generative LLMs, once fine-tuned to the target agricultural data domain, can generate a large amount of training data and labels to construct an augmented training set that closely resembles the original data distribution \citep{dai2023chataug}. Moreover, text-based generation models can synthesize images \citep{stablediffusion} and videos \citep{ho2022imagenvideo} that fit specific text descriptions that describe a specific scene. They can be used to generate customized agricultural visual data to fine-tune advanced computer vision algorithms.

\smallskip
\noindent\textbf{Multi-modal data editing and manipulation}. Agricultural data 
 show a wide range of variation due to changes in season and weather. It is extremely difficult to collect data that capture all the variations. Meanwhile, generative AI models show extraordinary abilities to edit and manipulate data with various attributes. They can be used to create variations of the original data for certain characteristics. Previous work on face editing show promising results such as changing a person's hair color and facial expression \citep{shen2020interpreting}.  
 In terms of image weather domain translation, generative adversarial networks (GAN) have demonstrated great potential in transferring \citep{gatys2016styletransfer,CycleGAN2017} ground-level images into photorealistic synthetic images under different weather conditions \citep{li2021weather,hwang2022weathergan} and different extreme climate events \citep{schmidt2022climategan}. Recently, diffusion-based image synthesis models such as SDEdit \citep{meng2022sdedit} also show their powerful image editing capabilities without the requirement of task-specific training. Corresponding to LLMs, large vision models are also developed, such as SAM for image segmentation \citep{kirillov2023segany}. 
 In the context of agricultural imagery, for example, generative AI models can change the time from day to night \citep{Punnappurath_2022day2night} and the weather from sunny to rainy. These variations can help in training more robust models for critical agriculture tasks.

\subsection{LLMs for Multi-Task Scheduling and Compositing}

The visual understanding of agriculture data usually consists of multiple sub-tasks. For example, for the main task of counting the number of healthy chickens on a poultry farm using cameras, sub-tasks such as object detection \citep{redmon2016yolo}, instance segmentation \citep{he2017maskrcnn}, object tracking \citep{wojke2017tracking}, and image classification \citep{he2016resnet} are needed. It requires domain knowledge in agriculture as well as in computer vision and machine learning to design and conduct such a workflow. It is simply not possible to scale. 
Taking another example, plant root phenotyping \citep{lu2021simultaneous, lu20213d, lu20223d} would involve 3D root structure modeling, root branch counting, length and angle estimation, biomass measurement, distribution pattern, etc, each of which involves extensive efforts. 

\begin{figure*}[t] 
\centering
\includegraphics[width=\textwidth]{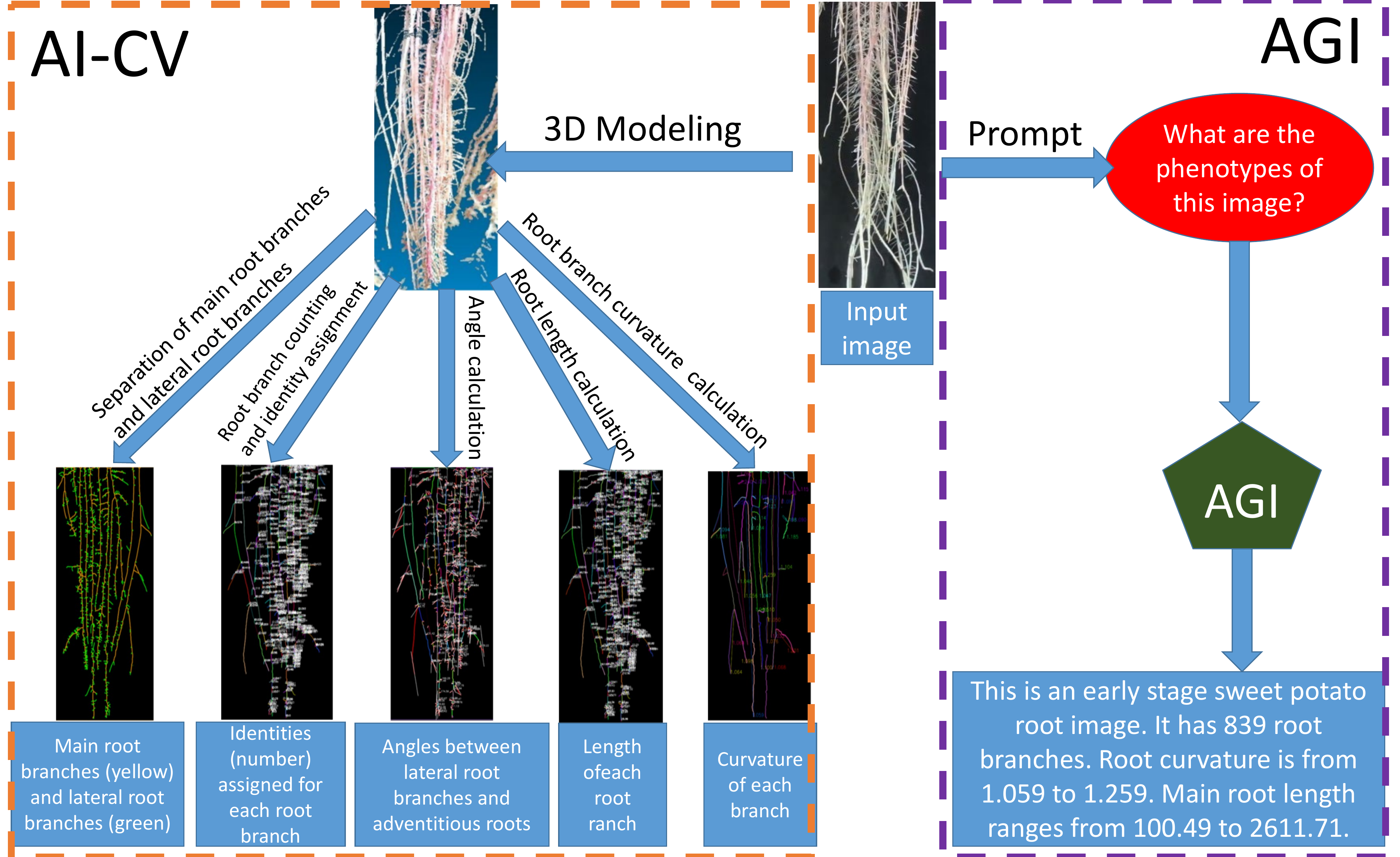} 
\caption{Plant root phenotyping based on AI-CV (computer vision) methodologies versus AGI. AI-CV requires extensive specific CV knowledge to process each task, while AGI has the potential to directly provide needed phenotype information with the human prompt. } 
\label{agiroot}
\end{figure*}

With the semantic understanding and reasoning ability of LLMs and their multimodal variants, it is now possible to make cutting-edge specialized vision and language models more accessible to important agriculture infrastructures such as farms. A recent study shows that ChatGPT is able to understand a request in natural language, extract useful visual and textual information, choose relevant vision and language tasks, and interpret the results back to humans \citep{shen2023hugginggpt}. LLMs can composite efficient and effective workflows from multiple vision and language tasks to perform critical agriculture tasks, all from simple and intuitive natural language instruction and without human intervention along the way. With the development of LLMs, it is expected that with the language input, the phenotype information could be directly provided to the agriculture researchers and breeders, as in Fig. \ref{agiroot}.

\subsection{Few-Shot Learning}

Unlike human intelligence which can infer new knowledge based on limited or zero experiences, most machine learning approaches, such as the traditional supervised learning techniques, often require thousands of labeled examples in order to learn simple patterns. To bridge this gap between human intelligence and AGI, few-shot learning (FSL)~\citep{fei2006one,fink2004object} has been proposed. In particular, FSL approaches focus on ``meta-learning'', that is, learning a model that can generalize to novel tasks specified at inference-time \citep{Vinyals2016}. In other words, FSL aims to emulate the remarkable generalization ability of human intelligence, by quickly adapting to new tasks that contain insufficient samples. This is very helpful for scenarios where data collection and labeling is difficult and expensive, such as agricultural applications. 

Depending on the sample availability, FSL can be divided into three categories: Few-Shot (two to five samples per class), One-Shot (a single sample per class), and Zero-Shot (classifying unseen classes without any training samples). FSL has shown promising results in various domains and applications, including computer vision~\citep{benaim2018one,shyam2017attentive,kozerawski2018clear,santoro2016meta} and natural language processing~\citep{ye2021crossfit,bansal2019learning}. Based on how to leverage the prior knowledge, existing FSL works can be categorized into three folds: 1) Data Augmentation. Some methods try to directly augment the training dataset by generating new samples from existing ones, transforming samples from a weakly labeled or unlabeled dataset, or transforming samples from a similar dataset. These augmentation processes are usually guided by hand-crafted rules, such as translation~\citep{benaim2018one}, flipping~\citep{shyam2017attentive}, reflection~\citep{kozerawski2018clear}, and rotation~\citep{santoro2016meta} operation for image data, or learning a set of transformations or models to conduct data augmentation~\citep{miller2000learning,schwartz2018delta,kwitt2016one}. 2) Hypothesis constraint. To approximate the ground-truth distribution, models may need to determine a hypothesis space that minimizes the distance between the optimal prediction and the ground-truth distribution. As a larger hypothesis space requires more data samples to explore, some FSL methods try to constrain the searching space of the hypothesis via jointly learning multiple related tasks with shared model parameters~\citep{motiian2017few,zhang2018fine,zhang2021survey} or restricting the form of distribution from prior knowledge~\citep{joshi2018extending,lake2015human,salakhutdinov2012one}. 3) Prior knowledge. When the training dataset is not large enough, FSL methods may leverage the prior knowledge to intervene in the searching strategy of the parameters, either by providing a good initialization of the parameters~\citep{azadi2018multi,caelles2017one,hoffman2013one} or directly learning an optimizer to output search steps, such as changing the search direction or step size~\citep{andrychowicz2016learning,ravi2017optimization}.

Although few-shot learning approaches have received a significant amount of attention over the past decade \citep{JimenezRezende2016, Sung2018, Kang2019, Yan2019, Yoon2020, Fan2020}, they have rarely rivaled the popularity of their fully-supervised brethren. Additionally, most such approaches have limited generalizability, applying only to a single type of task, such as image classification \citep{Vinyals2016}, object detection \citep{Kang2019, Yan2019}, or multi-object tracking \citep{Yoon2020}. An approach designed for, say, object classification cannot trivially be transferred to object detection, even though the two problems are related. One of the more surprising attributes of LLMs is that they exhibit a natural aptitude for few-shot learning \citep{Brown2020a}. Given the rapid progress of LLMs towards ever-more-general problem solving, it is not unreasonable to assume that we are entering a future where the current, data-hungry supervised learning paradigm is obsolete. Currently, a significant challenge for agricultural deep learning is a lack of annotated data, leading to a variety of attempts to reduce the need for it \citep{Adke2022, Akiva2020, Bellocchio2019, Bollis2020, Ghosal2019, Giuffrida, Marino2019, Petti2022, Ubbens2020}. Suitable data for a particular crop may not exist at all, may not be annotated properly, or may not be publicly available.

Advanced, multi-modal LLMs have the potential to solve these problems once and for all, but significant questions remain. Next-generation LLMs are only just beginning to integrate visual inputs \citep{OpenAI2023}. For its part, OpenAI is fairly tight-lipped about what exactly GPT-4 is capable of in this domain, and we likely won't know for sure until it is tested rigorously and independently. One clear risk is that LLMs trained on internet-scale data will fail to generalize well to the agricultural vision domain because agricultural images tend to be very different from the majority of pictures on the internet. In this situation, we would be forced to rely on the model's innate few-shot learning capabilities, which may or may not be up to the task. Additionally, there are many tasks of interest to the agricultural robotics community that require taking more obscure modalities (eg. point clouds or hyperspectral cubes) as inputs, and/or have outputs that cannot be expressed as text. Models such as GPT-4 would likely need to be extended at the architectural level in order to support such use cases. Given the fact that state-of-the-art LLMs are mostly proprietary, shrouded in secrecy, and impossible to train without millions of dollars invested in compute capacity, the research community might be waiting quite a while for these features.

If we take the optimistic view that near-future LLMs will be capable of at least some agricultural vision tasks, what would that enable? Already, thanks to the proliferation of tools like YOLOv5 \citep{Jocher2022}, simple computer vision tasks such as object detection can be accomplished by non-experts with little specialized training. Clearly, a reliable few-shot approach to agricultural computer vision would only make things easier by freeing us from the tyranny of data annotation, one of the last major pain points of computer vision workflows. Projects that would previously be difficult or slow due to the need to collect and annotate data could be deployed significantly faster. In addition, because of their language processing capabilities, vision-capable LLMs would be ideal for managing image-associated metadata, perhaps gaining the ability to derive textual metadata automatically based on an image input. Overall, LLMs stand to substantially accelerate the cycle of innovation within agricultural computer vision, and will (hopefully) usher these technologies into widespread use amongst farmers.

\subsection{Domain Adaption}

Though AGI has made breakthroughs in various application fields for its impressive performance in prediction and generalization, it unavoidably requires large amounts of labeled data which may not be realistic in real-world scenarios due to the considerable time and expensive labor forces. Unsupervised domain adaptation (UDA) provides a promising way to address this issue: it aims to adapt the models well-trained on one domain (source domain) to another domain (target domain) with different distributional characteristics, without the need for labeled data in the target domain~\citep{wang2018deep,farahani2021brief}. The key step of UDA is to find the discriminant and domain-invariant features that can link the labeled source domain and the unlabeled target domain in a latent space. In the past decades, many UDA methods have been developed and evaluated~\citep{ganin2015unsupervised,liu2019transferable,long2018conditional,shu2018dirt,zhang2019bridging,zhu2021cross,zhu2022self}. In addition to UDA, some other practical domain adaptation problems have also been extensively studied in the literature, such as universal domain adaptation~\citep{saito2020universal,zhu2021self}, class-imbalanced domain adaptation~\citep{shi2022pairwise}, few-shot domain adaptation~\citep{motiian2017few}, open-set domain adaptation~\citep{busto2017open}, and domain generalization~\citep{zhou2021domain,Zhu2022crossmatch}. 

In recent years, domain adaptation has been applied to some agricultural applications, such as plant classification~\citep{gogoll2020unsupervised}, leaf counting~\citep{valerio2019leaf}, corn yield prediction~\citep{ma2021multi}, agricultural land extraction~\citep{zhang2022unsupervised}, and land cover segmentation~\citep{bengana2020improving}. For instance, an unsupervised domain adaptation approach is proposed to transfer plant classification systems to new field environments, crops, and robots~\citep{gogoll2020unsupervised}. An adversarial domain adaptation method is proposed for agricultural land extraction using remote sensing images~\citep{zhang2022unsupervised}. In general, most previous work primarily relies on convolutional neural networks (CNNs) to learn domain-invariant representations~\citep{wang2019transferable,xu2018unsupervised} among different domains. With the surge of Vision Transformers (ViT) for vision tasks, a few ViT-based studies have been developed and demonstrated significantly superior transferability over their CNN-based counterparts~\citep{yang2023tvt,sun2022safe,xu2021cdtrans}. Moreover, the latest development on large multimodal models such as GPT-4 would lead to new formulation and modeling of domain adaptation, especially for certain application domains like agriculture. In traditional domain adaptation settings, the source domain usually offers auxiliary knowledge that is limited to a narrow and specific scenario. However, large multimodal models have already encoded rich commonsense knowledge, and they also have remarkable reasoning capabilities. Thus, the problem of transferring knowledge from large multimodal models to target domains should be carefully reformulated. Meanwhile, new methodologies on transfer learning, cross-domain representation learning, and visual/textual prompt-based knowledge transfer should be further developed.

\section{AGI for Agricultural Knowledge Graphs}
\label{AGI-KG}

Represented as labeled directed multigraphs, knowledge graphs (KGs) are a novel paradigm for the representation, retrieval, and integration of data from various data sources across domains \citep{bizer2011linked,janowicz2022know}. KGs have emerged as a promising solution for various academic and industry challenges \citep{noy2019industry}. In the following, we discuss several potential aspects where knowledge graphs technologies and AGI can work together and achieve a more intelligent agriculture system.

\textbf{Accurate and Interpretable AGI for Agriculture Tasks. }
Although the recent foundation models show promising performances on various language, vision, and robotic tasks \citep{bommasani2021opportunities}, they are also criticized for generating inaccurate and misleading results \citep{touvron2023llama,huang2023language}. Knowledge graphs have been shown to be particularly useful to guide the predictions of language models \citep{yasunaga2022deep} on different tasks and help ground the generated answers on a particular subgraph \citep{sun2018open} to ensure the accuracy of generated answers and improve the model interpretations. This capability is particularly useful for
many real-world tasks in which accurate and interpretable predictions are critical such as tasks in medical and agriculture domains. With the contextual information provided by knowledge graphs, AGI can be better used for agriculture tasks such as crop monitoring, crop yield forecasting, fertilizing, and so on.

\textbf{Text-Knowledge Fusion for Agriculture Tasks. }
With the recent development of large language models (LLMs) which are generally treated as the major driving force behind AGI, knowledge graphs are commonly treated as an additional data source that can complement text data
for language model pretraining or finetuning to significantly boost their performance on various downstream tasks, especially difficult question answering tasks that require complex reasoning \citep{sun2018open,mai2020se,xiong2020pretrained,agarwal2021knowledge,mai2021geographic,zhang2022greaselm,yasunaga2022deep}. In the agriculture domain, AgriBERT \citep{rezayi2022agribert} has shown that augmenting text data (i.e., question answering pairs) with additional knowledge from an agriculture knowledge graph (i.e., FoodOn knowledge graph) can significantly boost model performance on various NLP tasks in the food and agriculture domain.

\textbf{AGI for Agriculture Knowledge Graph Construction. }
The recent developments of foundation models foster the usage of AGI on knowledge graph construction. Instead of performing knowledge triplet extraction in a supervised manner~\citep{christopoulou2021distantly} or by using seed samples~\citep{chia2022relationprompt}, Mai et al.~\citep{mai2022towards} demonstrate that by using simple prompts, GPT-3 can accurately recognize place names from unstructured text and outperforms various state-of-the-art fully supervised toponym recognition models. This shows the potentials of AGI for entity recognition and entity linking which are two critical tasks related to knowledge graphs. Recently, GraphGPT\footnote{\url{https://github.com/varunshenoy/GraphGPT}} has been developed as a GPT-3 based knowledge graph generation tool that can update a given knowledge graph or create a new knowledge graph given natural language inputs. Most widly used knowledge graphs, either general-purposed KGs such as \textit{DBpedia} \citep{auer2007dbpedia}, and \textit{Wikidata} \citep{vrandevcic2012wikidata}, or domain-specific KGs such as \textit{FoodOn} \citep{dooley2018foodon}, \textit{GeoNames}\footnote{\url{https://www.geonames.org/}} \citep{ahlers2013geonames} \textit{LinkedGeoData} \citep{auer2009linkedgeodata}, 
\textit{GNIS-LD} \citep{regalia2018gnis}, 
and \textit{KnowWhereGraph}\footnote{\url{https://knowwheregraph.org/}} \citep{janowicz2021knowwheregraph,janowicz2022know}, are constructed manually, or semi-automatically. AGI-based automatic knowledge graph construction is a promising research direction, especially for domains where there are fewer knowledge engineers available such as the agriculture domain. The automatically constructed agriculture knowledge graph can provide additional contextual information and facilitate various agriculture decision-making processes.

\textbf{Complex Knowledge Reasoning for Agriculture Tasks. }
Many agriculture tasks require complex reasoning~\citep{bhuyan2022systematic}. For example, given a UAV image of a cropland of soybean, an agricultural scientist or a farmer expect an AGI can 1) first detect whether there are some abnormal symptoms given the appearance of soybean leaves (e.g., water-soaked spots on leaves); then 2) determine the name of this problem (e.g., bacterial blight); next, 3) determine the causes of this disease (e.g., pseudomonas syringae); and finally, 4) determine how to treat this disease (e.g., spray bactericides). Currently, one popular way to enable a large language model (LLM) to do complex reasoning is the usage of chain-of-thought prompting \citep{wei2022chain} where a few chains of thought demonstrations are provided as exemplars in prompting to explicitly force the LLM to conduct complex reasoning. 
The idea is to allow LLM to decompose a multi-step reasoning problem into intermediate steps that are solved individually, instead of solving an entire multi-hop reasoning problem in a single forward pass. However, such an idea is to reason in an implicit manner which is prone to errors and hard to debug. By explicitly encoding the relationships among crop symptoms, diseases, bacteria, and treatments into an agriculture knowledge graph, we can ask the LLM to follow the path in this graph and conduct explicit multi-hop reasoning just as the way Neural Symbolic Machine adopts \citep{liang2017neural}.

\textbf{Applications of Knowledge Graphs to Agricultural AGI.} Knowledge graphs describe the properties and relationships of real-world entities, providing benefits to a variety of agricultural applications~\citep{chen2019agrikg}, such as enhancing search engines, Q\&A systems, recommendation systems and content generation with greater intelligence.

\begin{itemize}
    \item Search Engines.  
    Agriculture search engines could better connect farmers and professionals to the information they need~\citep{ingram2019searching}. Traditional search engines work by crawling and indexing web content, accepting queries, and ranking results according to relevance~\citep{schutze2008introduction}. Due to the emerging general intelligence, the recent large language models (LLMs) such as ChatGPT and GPT-4~\citep{OpenAI2023} could be used to generate results for searches, potentially disrupting the search engine industry. The concept of knowledge graphs gained great popularity since its launch by Google's search engine, associated with the Knowledge Vault framework~\citep{dong2014knowledge} to build large-scale knowledge graphs. Knowledge graphs can complement LLMs by providing them with structured and reliable information.
    \item Q\&A Systems. Question Answering (Q\&A) aims to provide accurate and concise natural language answers in response to questions posed by users~\citep{zhu2021retrieving, jain2019agribot, mai2021geographic}. The recent chatbot-style language models, if trained with large amounts of agricultural data from various sources (e.g., weather patterns, and pest control), have the potential to answer questions related to various agricultural domains. Nevertheless, LLMs may generate content that lacks factual accuracy or omits important knowledge, where KGs could be applied to mitigate the problem. One strategy is to enhance language representation learning with knowledge entities~\citep{zhang2019ernie,xiong2020pretrained}. Another way is to directly learn knowledge graph embeddings and then integrate the learned entity and relation embeddings into the QA pipeline~\citep{huang2019knowledge, mai2019contextual, saxena2020improving}.
    \item Recommender Systems. Agricultural recommender systems can facilitate the dissemination of authentic information and help farmers make more informed decisions on how to optimize their production, e.g., suggesting which agricultural products to use, such as seeds, fertilizers, and pesticides~\citep{garanayak2021agricultural}. The recent conversational AI~\citep{sun2018conversational} could even interact with users to better help them find products/services. By utilizing the side information in KGs, it is possible to capture user preferences more accurately~\citep{guo2020survey}. Typical strategies of combining KGs include learning informative knowledge graph embeddings~\citep{wang2018dkn,liu2020explainable} and conducting path-based reasoning over KGs~\citep{xian2019reinforcement,ma2019jointly}.
    \item Content Generation. The advanced language processing capabilities of LLMs can simplify and speed up repetitive mental tasks of agricultural professionals, such as recording, report generation, and training materials generation\footnote{\url{https://agtecher.com/how-openai-and-chatgpt-can-be-used-in-agriculture/\#help-agri}}. The incorporation of KGs could make the content generation process more controllable~\citep{yu2022survey}. Besides knowledge graph embedding~\citep{zhou2018commonsense,cai2019transgcn} and path-based reasoning~\citep{liu2019knowledge}, it is also possible to apply graph neural networks to encode high-order information towards text generation~\citep{beck2018graph}.
\end{itemize}

 \section{AGI for Agricultural Robotics}
\label{AGI-Robotics}

By enhancing the decision-making process of robots, AGI can improve the efficiency of executing certain tasks, particularly in the agricultural sector where some tasks still rely heavily on manual labor \citep{xu2022review}. 
The application of AGI in agricultural robotics can enable robots to perform a range of tasks that currently require many single-task AI models to perform \citep{xu2022modular,iqbal2020development}. It also has the potential to revolutionize the design of agricultural robots by leveraging high-level function libraries and prompt engineering \citep{vemprala2023chatgpt}. 
With AGI, agricultural robots can optimize various tasks and minimize waste, making them a promising technology for precision agriculture. 

\begin{figure*}[t]
    \centering
    \includegraphics[width=\textwidth]{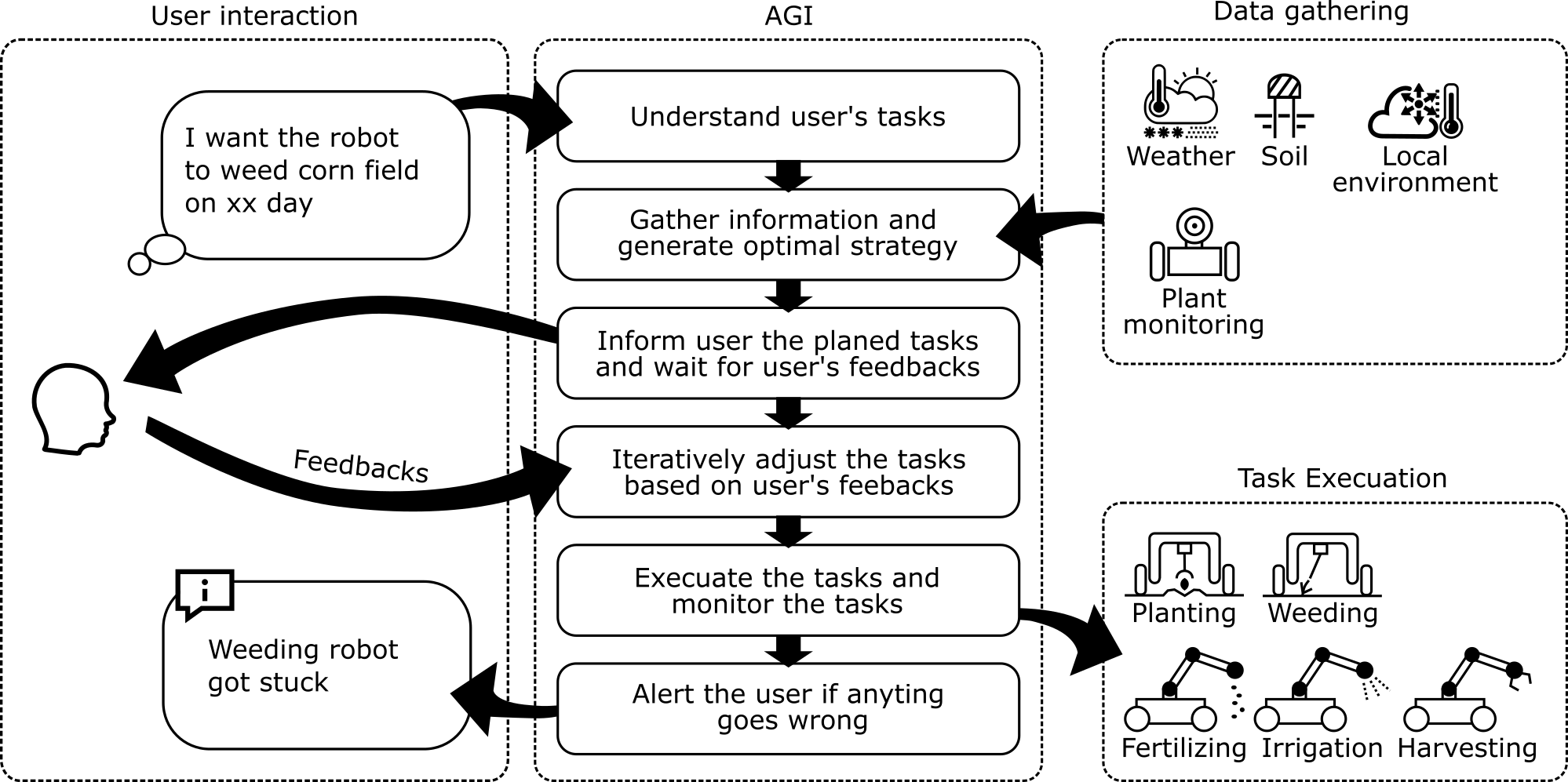}
    \caption{Future scenarios of enhanced human interaction with agricultural robots with the aid of AGI to perform farming operations more efficiently.}
    \label{fig:agi_robot}
\end{figure*}

AGI's ability to understand natural languages and reasoning can greatly improve the interaction between humans and robots, thereby lowering the technical barriers for farmers to use agricultural robots. Additionally, AGI's ability to comprehend images can enable robots to better understand their surroundings through computer vision, which includes identifying potential safety hazards. The future scenarios of interactions between humans and agricultural robots are illustrated in Fig \ref{fig:agi_robot}. These are just a few examples of how AGI can enhance the applications of agricultural robots. 

\begin{itemize}

\item \textbf{Human-robot interaction}. AGI can significantly simplify the interaction between humans and robots by allowing users to control robots using natural language, rather than typing out commands. This reduces barriers to usage for those without a technological background and saves costs on training certified operators. 

\item \textbf{Robot safety}. AGI's ability to detect safety hazards in real-time can significantly improve the safety of agricultural robots. For instance, if a robot detects a safety hazard, it can immediately stop its operation or alert the farmer, preventing potential accidents or injuries. The robots can also monitor the progress of agricultural tasks and identify potential damage to plants when command execution fails. Additionally, AGI can help robots learn from their experiences and improve their performance over time, reducing the likelihood of future safety hazards. 

\item \textbf{Human-robot collaboration}. AGI can enable robots to coordinate more effectively with humans in more complex tasks, such as harvesting. For example, a harvest aid robot can offer more efficient and responsive assistance to a human worker when it can understand the worker's intentions by recognizing their gestures and language, improving the overall efficiency of harvesting. Additionally, the robot can identify potential hazards and avoid harming human workers. 

\end{itemize}

AGI has several advantages, including the ability to integrate historical and real-time data from various sources such as weather, environment, soil, and plant data. This can help robots generate optimal strategies for specific agricultural tasks in real-time. The applications of AGI in agriculture are numerous, including enhancing the efficiency of the following tasks: 

\begin{itemize}
    \item  \textbf{Planting}. AGI can be used to help robots plant crops more efficiently by providing real-time data on soil quality, moisture levels, and other factors that affect seed growth. By using this data, robots can plant seeds at optimal depths and spacing, improving crop yields and reducing waste. 
    \item  \textbf{Weeding}. AGI can help robots differentiate between weeds and crops and eliminate weeds using targeted spraying, laser, or mechanical removal without damaging the crops, which can greatly reduce the usage of herbicides and their impacts on the environment. 
    \item \textbf{Fertilizing}. AGI can be used to detect the plant growth status and identify plants suffering from malnutrition. Combined with data on soil quality, robots can adjust the amount and timing of fertilizers to optimize crop yields and reduce waste. 
    \item \textbf{Harvesting}. AGI can help robots to detect the ripeness of the crops and determine the optimal harvesting timing, enabling robots to selectively harvest the ripe crops to ensure good quality. AGI is particularly useful to enable robots to perform fruit picking for certain crops that currently can only be harvested manually. Using AGI, robots can learn to recognize the specific features of each crop and adjust their picking strategy accordingly to improve harvesting efficiency. 
    \item \textbf{Irrigation}. AGI can utilize weather data from online sources, local environmental data and soil moisture from IoT devices, and plant growth from crop monitoring to optimize irrigation schedules. AGI can help robots precisely control the irrigation amount for each plant, reducing wasting water and improving crop yields. 
    \item \textbf{Crop scouting}. AGI can help robots to monitor the growth of the crop and identify potential issues in the early stage. By analyzing the crop data collected using sensors and cameras, AGI can monitor the physiological activities of individual plants, which are important information to make decisions for other agricultural tasks. AGI can also help robots detect problems such as water stress, nutrient deficiencies, pest infestations, and disease outbreaks, enabling farmers to take corrective action before crop yields are affected.  
    \item \textbf{Phenotyping}. AGI can be a great tool for plant breeders to quickly extract phenotypical traits from the large amount of data collected by robots. AGI can be used to automatically organize data, perform basic statistical analysis, visualize data, and generate reports, saving breeders time on data preparation and analytics. 
\end{itemize}

\section{A Few Case Studies
}
\label{AGI-application}

\subsection{AGI for Precision Farming and Phenomics}

The data generated from precision agriculture is typically in the form of raw data collected from various sensors, such as images, point clouds, spectra, and time-series measurements. In contrast, Large Language Models (LLMs) traditionally operate in the text domain. Therefore, the application of LLMs directly to Agriculture 4.0 may appear to be the naive coupling of two buzzwords. This may lead to initial skepticism regarding the potential usefulness of LLMs in the agriculture domain.

\begin{figure*}[h]
    \centering
    \includegraphics[width=0.75\textwidth]{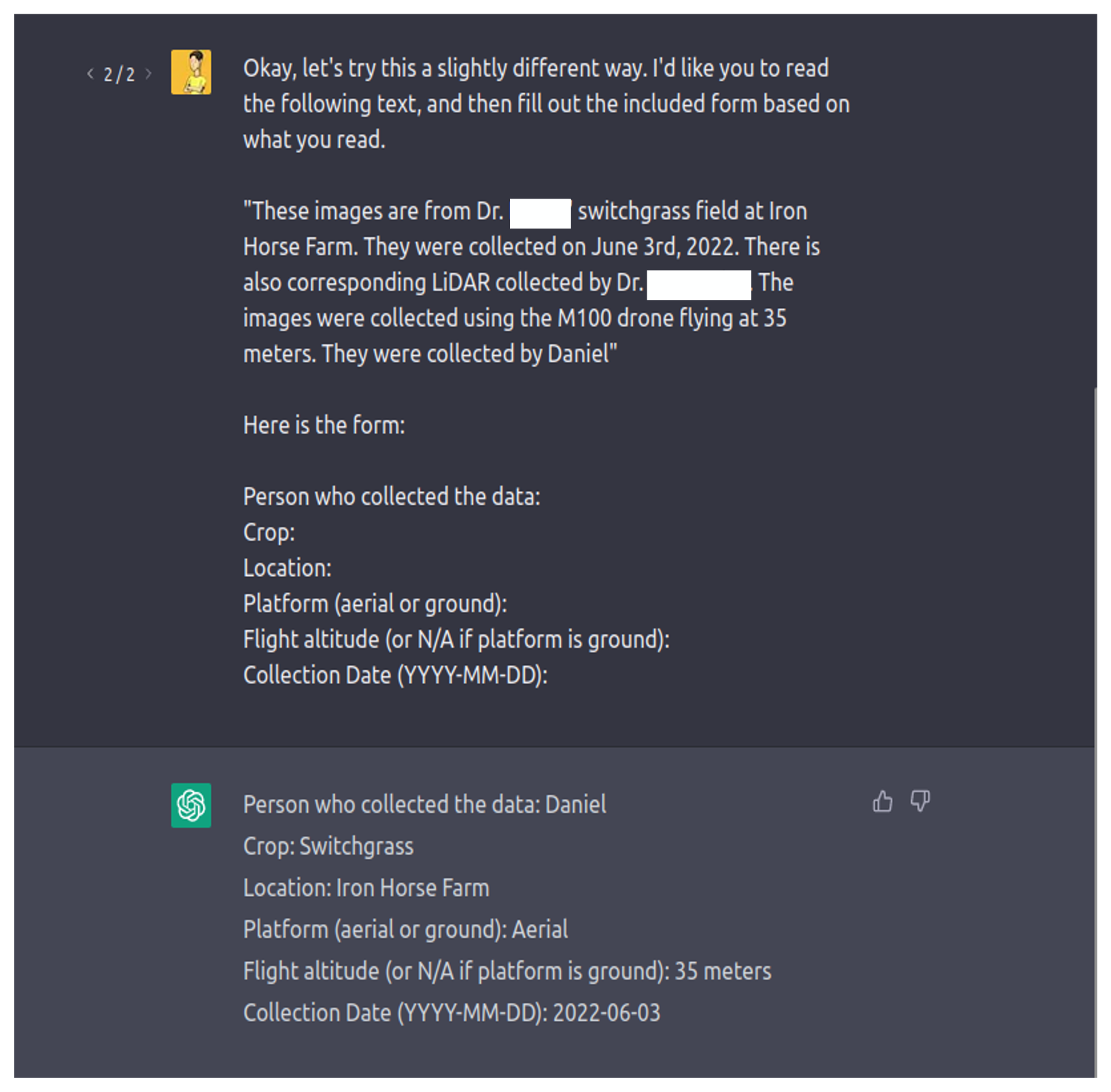}
    \caption{The current version of ChatGPT is extremely capable at converting unstructured text data to structured data.}
    \label{fig:chatgpt_form_filling}
\end{figure*}

In practice, however, many advances in agriculture are driven by experiments that produce large amounts of metadata, often expressed in natural language. For example, \citet{Papoutsoglou2020} attempts to formalize the structure of plant phenomics data, but the sheer complexity of their approach increases the barriers for potential users, as well as the risk of mistakes. LLMs have the potential to partially or fully automate tasks such as structuring unstructured metadata (Fig. \ref{fig:chatgpt_form_filling}), converting metadata from one format to another, and flagging possible mistakes during data collection. When coupled with voice recognition technology such as the Whisper APIs, LLMs could even assist plant breeders working in the field in collecting measurements without performing tedious and error-prone data entry. Additionally, \citet{Bubeck2023} suggests that next-generation LLMs will be extremely capable tools for data visualization, potentially making it much easier to glean useful insights from these text data. LLMs also have the ability to make searching large quantities of text data much more effective. Overall, LLMs have the potential to greatly assist researchers in gleaning useful insights from large quantities of phenotypic data.

\subsection {AGI for Precision Livestock}
\label{AGI-Poultry}

\begin{figure*}[h] 
\centering
\includegraphics[width=\textwidth]{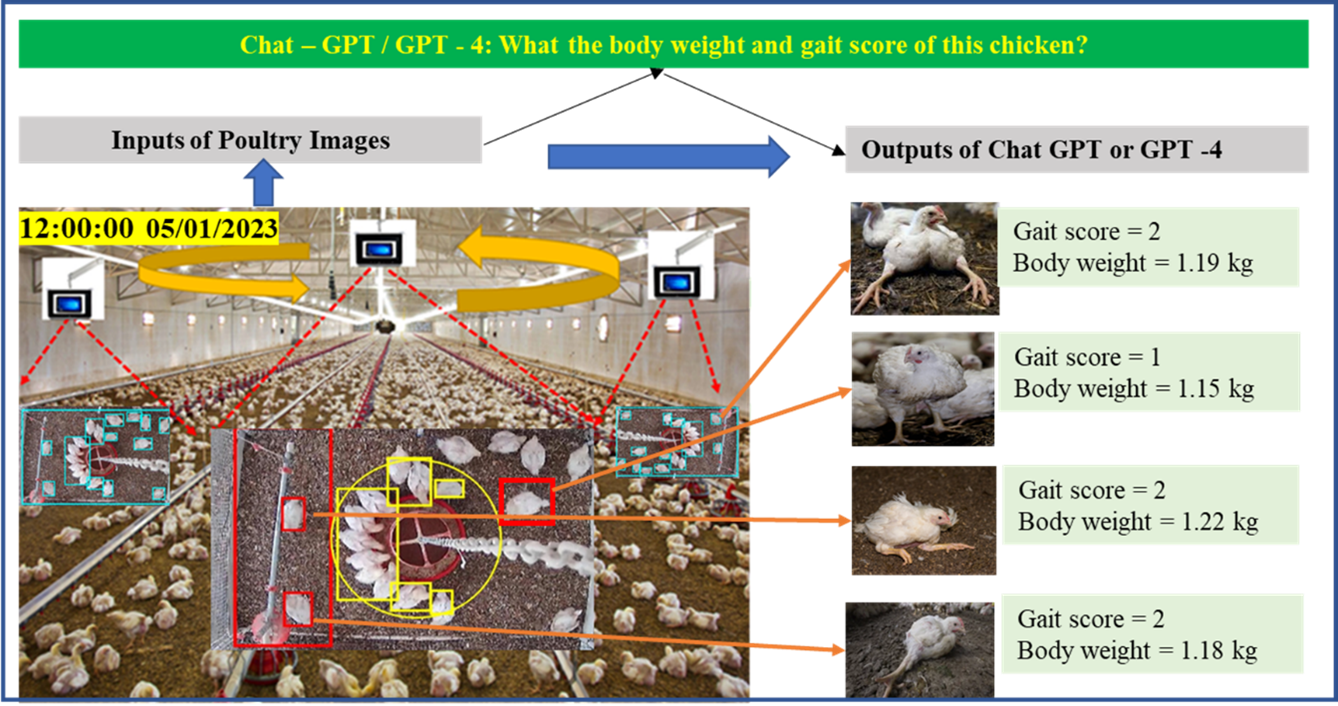} 
\caption{Generative AI system for automatic assessment of poultry body weight and welfare (gait score is walking ability indicator (0, healthy legs, 1, medium lameness, and 2- severe lameness similar to dead birds). } 
\label{poultry}
\end{figure*}

Poultry and egg production provide valuable and affordable protein sources for increasing world's population. In the past decades, the production efficiency of broiler chickens has been doubled in the US due to continuous innovations in animal breeding, nutrition management, environmental control, and diseases prevention \citep{chai2018mitigating} \citep{guo2022monitoring} \citep{subedi2023tracking} \citep{bist2023ammonia}. 
However, it is challenging to improve animal production efficiency and product quality under limited natural resources (e.g., fresh water and land), thus precision production is critical for addressing the issue. A key task of precision poultry production is monitoring animal behaviors for the evaluation of welfare and health \citep{dawkins2013search}\citep{fernandez2018real}\citep{ castro2023poultry}. Application of AGI such as GPT-4 for analyzing poultry videos and images requires innovation in the development of computer vision or machine vision system \citep{subedi2023tracking}. Vision transformer (ViT) has become one of the emerging methods for monitoring animal information and as a decision support tool in precision animal farming \citep{liu2022depthformer, lu2022recognition}. 
Previous works in poultry welfare analysis primarily focused on using computer vision techniques such as convolutional neural networks (CNNs) and image classification algorithms to detect birds’ feeding, drinking, or preening behaviors \citep{li2020developing}\citep{guo2022monitoring} \citep{subedi2023tracking1}. In recent years, advanced methods were reported in tracking chickens’ problematic behaviors such as pecking, mortality, floor distribution in broiler houses, and spatial distribution in cage-free hen houses \citep{guo2020machine}\citep{yang2022deep} \citep{yang2023deep}. However, monitoring poultry welfare and production with computer vision is challenging due to scarce of training poultry videos, labels of specific problematic welfare issues such as lameness in broilers and pecking damages in layers, and interpretation of analysis results to understandable language to poultry farmers. Previous methods are not applicable for poultry farming as growers usually don’t have knowledge of data analysis with ML or DL models. The emergence of ChatGPT or GPT-4 provides a new capacity for farmers as it would be not complicated to upload a chicken video or images into ChatGPT or GPT-4 as inputs and expect the outputs (e.g., specific poultry behavior or welfare) from ChatGPT or GPT-4 automatically (see Fig. \ref{poultry}).

\subsection {AGI for Agricultural Infrastructure} 
\label{AGI-infra}
Infrastructure management and facility training are of critical importance to the success and resilience in today's agricultural sector \citep{pawlak2020role}. Adequate infrastructure support, such as bridges, roads, irrigation, storage, IT, etc can greatly improve productivity, processing capability, and market values. Due to infrastructure modernization, the complexity for managing and training of this infrastructure grows exponentially, which makes traditional approaches infeasible. In recent years, we have witnessed the potential of AI and AGI tools for efficient processing and management of complex systems, which can extend to the agricultural field \citep{alreshidi2019smart, mclennon2021regenerative, bhat2021big, de2019scalable}. In this section, we will summarize and envision the role of AGI in agricultural infrastructure management and education.  

\subsubsection{AGI for Efficient Infrastructure Management } 
Reliable infrastructure is one of the key factors for ranchers and farmers to expand operations and deliver better products \citep{mykhailichenko2021competitive}. 
Obviously, such infrastructure has the potential to transform and modernize traditional agriculture and farming into a more efficient, sustainable, and dynamic farming system. It is a critical sector in terms of economic growth and development and sustainable food production. Specifically, infrastructure management includes agricultural roads, land development,  and other transportation infrastructure; water and irrigation, information systems, agricultural equipment, machinery, building infrastructure (some examples include greenhouses, warehouses, small industries connected with agricultural production, silos, tanks, etc.), energy production for/from agriculture, institutional infrastructure (agricultural research, extension \& education technology, information and communication services, financial services, marketing, etc.), and agricultural waste management systems, etc. \citep{infra}. 

Management of these infrastructures, however, is a challenging task. It requires coordination, proper allocation, and most importantly, efficient utilization. Existing approaches rely mostly on human intervention and scheduling, which cannot keep up with the scale of complexity. Therefore, future infrastructure management will rely on advanced ML tools \citep{javaid2023understanding}, such as AGI, to remedy this issue. AGI can deliver the following innovative infrastructure management tasks. 
\begin{itemize}
    \item \textbf{Predictive Maintenance.} AGI can be used to analyze data from sensors and other sources to predict when infrastructure components such as irrigation systems, farm machinery, and storage facilities are likely to require maintenance or repairs \citep{arumugam2022towards, khorsheed2021integrated}. This can help farmers and infrastructure managers take preventive measures, reducing the risk of equipment failure and downtime.
    \item \textbf{Resource Optimization.} AGI can help farmers and ranchers optimize the use of resources such as irrigation, fertilizer, and pesticides. By analyzing data from sensors and other sources, machine learning tools can help farmers determine the optimal timing and amount of resource application, reducing waste and increasing efficiency \citep{dey2021blockchain}. Besides, these tools can also be efficiently coordinated for better utilization in terms of capital cost or time-sharing values \citep{maraseni2021carbon, abbasi2022digitization}. 
 
    \item \textbf{Infrastructure Design and Planning.} AGI can be used to analyze data from weather sensors, soil sensors, and other sources to inform the planning and design of agricultural infrastructure such as irrigation systems, drainage systems,  storage facilities, and other sensors \citep{lutz2022applications, abbasi2022digitization}. This can help infrastructure managers optimize infrastructure design and improve facility efficiency.
\end{itemize}

As one example, poultry infrastructure shown in Fig. \ref{poultry} can be accelerated by AGI towards more efficient management. Modern poultry house equips with advanced vision sensors (such as RGB and infrared cameras),
fans, air quality monitoring devices, etc. With AGI data analytics and extraction, device maintenance will be performed more precisely, such that downtime is minimized. On the other hand, AGI can help optimize when and where to collect data from whom, for better device resource optimization. 

\subsubsection{AGI for Infrastructure Training and Education}
AGI can also help develop tools that can facilitate agriculture training and education. The diverse types of equipment/devices make training and education very challenging, time-consuming, and hard to generalize. Recently, the development of AGI can relieve such challenges. For example, the proliferation of learning tools has proven to be very effective in education and its feedback evaluation. In this line of research, we expect to see innovative tools, based on the principle of AGI, that can be utilized in the loop of infrastructure training and education.

AGI can help infrastructural training and education in the following aspects. 

\begin{itemize}
    \item \textbf{Personalized Training and Education.} AGI can be used to analyze trainee data and provide personalized training recommendations, allowing trainees to focus on areas where they need the most improvement \citep{chen2020application,zhang2020understanding}. Besides, depending on trainee's background, a customized learning pathway can be designed, which aims for better learning efficiency. 
    \item \textbf{Predictive Analytics and Feedback.} AGI algorithms can be used to analyze data from training sessions to identify trends and predict training outcomes \citep{alam2021should}. It is especially important since more training and education session  now move to online and the feedback are not easily sensible from the instructor's perspective. With advanced analytical tool provided by ML, it can help trainers optimize training programs and improve training effectiveness \citep{10.1007/978-981-19-8825-7_11, ayouni2021new, liu2023perspectives, zhai2020applying}. 
    \item \textbf{Continuous Learning.} AGI can be used to create continuous learning programs, providing farmers and ranchers with ongoing training and support even after the initial training program has ended. This can be very helpful as farmers and ranchers can receive life-long learning based on their progress, the development of new equipment, or their farming type \citep{marques2020teaching}. Compared with current approaches, this new method will be very effective, especially combining personalized learning and predictive feedback. 
\end{itemize}

\section{Conclusion}
\label{AGI-Conclusion}

AGI has great potential to revolutionize agricultural and food systems, 
including precision farming for crops and livestock, 
agriculture image processing, NLP for agriculture problem understanding and answering, 
agricultural robotics, agriculture and food knowledge graph, and agricultural infrastructure. AGI can provide farmers and agricultural professionals with valuable insights and advice on how to improve their productivity and sustainability, while also addressing the challenges facing the industry, such as climate change, food security, and rural development. AGI could have a significant impact on farm automation, where robots and drones equipped with advanced computer vision and machine learning capabilities could perform a range of tasks, from planting and harvesting to monitoring crops and livestock. With its ability to understand natural language, images, and generate human-like responses and enhance robotics capabilities, AGI would be an innovative tool for advancing agriculture and improving the livelihoods of people involved in this vital sector.

This article mainly summarizes the possible applications and effects of AGI in the domain of agriculture. However, the potential applications of AGI in agriculture are virtually limitless, and the industry is just beginning to scratch the surface of what is possible. Some of the other potential applications of AGI in agriculture include precision farming, where advanced algorithms can analyze data from sensors and other sources to optimize crop yields, reduce waste, and minimize environmental impact. AGI can also be used in developing new crop varieties through computational breeding, which could speed up the process of developing more resilient, high-yield crops. The potential applications of AGI in agriculture are vast and varied, and there is no doubt that as the technology continues to evolve, we will see many exciting new developments in this field.

\bibliographystyle{cas-model2-names}


\end{document}